\documentclass[conference]{IEEEtran}
\IEEEoverridecommandlockouts
\usepackage{cite}
\usepackage{amsmath,amssymb,amsfonts}
\usepackage{algorithmic}
\usepackage{graphicx}
\usepackage{textcomp}
\usepackage{xcolor}
\usepackage[utf8]{inputenc}
\usepackage{subcaption}
\graphicspath{ {./images/} }
\usepackage{array}
\usepackage{multirow}
\usepackage{makecell}

\def\BibTeX{{\rm B\kern-.05em{\sc i\kern-.025em b}\kern-.08em
    T\kern-.1667em\lower.7ex\hbox{E}\kern-.125emX}}
\begin{document}

\title{%
Integrating High-Resolution Tactile Sensing into Grasp Stability Prediction
}
\author{Lachlan Chumbley$^{1}$, Morris Gu$^{1}$, Rhys Newbury$^{1}$, J\"urgen Leitner$^{2}$ and Akansel Cosgun$^{1}$ \\
$^{1}$Monash University, Australia\\$^{2}$LYRO Robotics, Australia
\\ \{lachlan.chumbley,morris.gu,rhys.newbury,akansel.cosgun\}@monash.edu, juxi@lyro.io}


\maketitle

\begin{abstract}


We investigate how high-resolution tactile sensors can be utilized in combination with vision and depth sensing, to improve grasp stability prediction. Recent advances in simulating high-resolution tactile sensing, in particular the TACTO simulator, enabled us to evaluate how neural networks can be trained with a combination of sensing modalities. With the large amounts of data needed to train large neural networks, robotic simulators provide a fast way to automate the data collection process. We expand on the existing work through an ablation study and an increased set of objects taken from the YCB benchmark set. Our results indicate that while the combination of vision, depth, and tactile sensing provides the best prediction results on known objects, the network fails to generalize to unknown objects. Our work also addresses existing issues with robotic grasping in tactile simulation and how to overcome them.


\end{abstract}

\section{Introduction}
Advancements in grasping and manipulation abilities are one of the major factors that will allow robots to move outside the world of manufacturing. Moving from structured environments to dynamic, complex human environments will allow robots to assist in homes, offices, and hospitals. This transition to operate in these human-centered environments will require high levels of dexterity, intelligence and versatility. As such, robots will require human-like abilities in grasping and manipulation.

\begin{figure}[t!]
    \centering
    \includegraphics[width=1\linewidth, trim={0 0 0cm 0},clip]{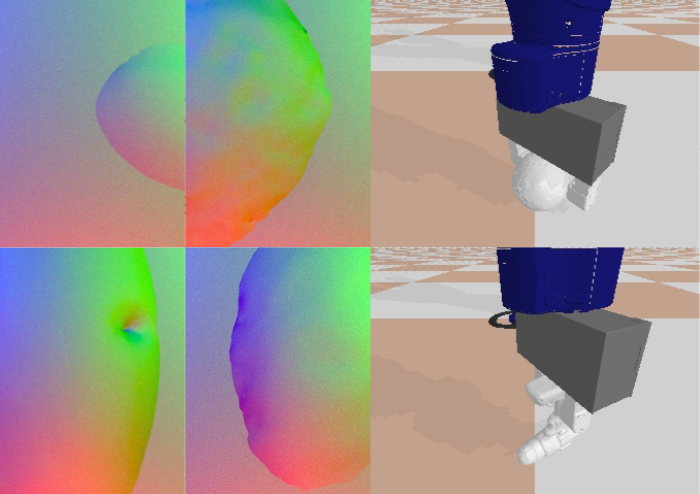}
    \caption{Two examples of input to our approach for a successful grasp. Each row of three images, shows the output of the TACTO simulator, for both left (left image) and right gripper fingers (middle image), with the image on the right showing a camera view of the scene. Our network can successfully predict grasp success using this data. The intensity of the pixel roughly represents the depth of displacement for the gel inside the DIGIT sensor, where the darker the pixel, the greater the depth. 
    }
    \label{fig:intro}
\end{figure}

Current approaches to robotic grasping often utilize a single sensing modality, commonly RGB-D cameras~\cite{pas2017grasp}. It remains uncommon to use tactile sensors for grasping, especially in conjunction with another sensing modality such as vision. In contrast, humans heavily rely on the sense of touch when they are manipulating objects~\cite{johansson2009}. This discrepancy is due to the difficulty in creating robotic end-effectors that are as capable as human hands, owing to a lack of sensors and actuators equivalent in size, precision, and efficiency to human skin and muscles~\cite{billard2019trends}.


A major challenge with machine learning in robotics is insufficient amounts of training data due to the limitation of the data collection speed with real hardware \cite{Du_2020}. To combat this, simulated environments are often used to collect datasets much larger and faster than previously obtained using a real robot~\cite{mahler2017dexnet,depierre2018jacquard,fang2020graspnet}. However, due to the difficulty of simulating vision-based tactile sensing, it has been largely excluded from these simulated grasping experiments.
The recent development of TACTO~\cite{wang2020tacto}, an open-source simulator for high-resolution vision-based tactile sensors, has bridged this gap and allowed for the large-scale simulated grasp sampling required to train a deep neural network. \cite{wang2020tacto} demonstrates the feasibility of the simulator with an example scenario where a network is trained to predict the grasp success of a single, rectangular object given visual tactile readings, however, to our knowledge, extensive experiments with TACTO has not been conducted in the literature.

We pose grasp stability prediction as a supervised learning task and use Convolutional Neural Networks (CNN) for function approximation. The input to the network is a set of images coming from multiple sensors and the output is a binary label indicating whether the grasp would be successful if the object is lifted. In this work, we use the TACTO simulator to train a network for predicting grasp success for a given grasp from tactile, visual and depth sensor data. Our grasping experiments are conducted using a subset of the YCB object set~\cite{calli2015ycb}, a well-known benchmark object set in robotic grasping research, extending previous work using only a single object~\cite{wang2020tacto}.

The primary contributions of this paper are:
\begin{enumerate}
\item Ablation studies with different combinations of sensing modalities including RGB and depth images from a side camera and tactile (DIGIT) sensor images.
\item An investigation into validating the TACTO simulation of tactile sensors for robotic grasping.
\item A data filtering process for robotic grasping dataset collection, particularly for handling tactile information.
\end{enumerate}


\section{Related Work}
\label{sec:related_work}



\subsection{Tactile Sensors}

Developing robotic skins is an active research area~\cite{hughes2018}, however, these sensors remain niche, hence are not commonplace. An alternative to robotic skins are high-resolution vision-based tactile sensors~\cite{yuan2017gelsight,Dong_2017,donlon2018gelslim,taylor2021gelslim30,yamaguchi2017fingervision,lambeta2020digit,omnitact2020}, which provide a high spatial resolution by typically outputting an image that encodes the deformation on the contact surface. 


Data-driven methods alongside the development of sensors, particularly depth sensors, have allowed deep learning training to be performed with multiple modalities, significantly improving the robotic grasping capabilities~\cite{pas2017grasp}. Recently, the development of high-resolution tactile sensors has allowed for tactile information to be incorporated into the training, improving performance in both grasping~\cite{calandra2017feeling,Calandra_2018} and manipulation~\cite{tian2019manipulation,hogan2020dexterity,dong2021tactilerl}.

Early work in tactile sensors largely focused on measuring force and torque applied to the end-effector or the sensor's pressure distribution over the sensor~\cite{YOUSEF2011171}. 

There are at least eight main tactile sensor types, including ~\cite{KAPPASSOV2015195}; piezoresistive~\cite{koivatactile2013}, capacitive~\cite{koivatactile2013}, piezoelectric~\cite{schmitz2011tactile}, quantum tunnel effect~\cite{strohmayr2013artificial}, barometric measurements based~\cite{fishel2008robust}, multi-modal~\cite{hasegawa2010intelligent}, structure-borne sound~\cite{liangting2012seashell} and vision based~\cite{yuan2017gelsight,Dong_2017,donlon2018gelslim,taylor2021gelslim30,yamaguchi2017fingervision,lambeta2020digit}.

High-resolution vision-based tactile sensors such as GelSight~\cite{yuan2017gelsight,Dong_2017}, Gelslim~\cite{donlon2018gelslim,taylor2021gelslim30},
FingerVision~\cite{yamaguchi2017fingervision}, OmniTact~\cite{omnitact2020} and DIGIT~\cite{lambeta2020digit} have been applied to robotic manipulation~\cite{tian2019manipulation,lee2019making,hogan2020dexterity,hogan2020tactile,dong2021tactilerl} and slip detection~\cite{yuan2015gelsiteshear,Dong_2017} with some success. The vision-based sensors have a high spatial resolution and ability to provide a better generalization to objects of different geometry compared to Force/Torque sensors~\cite{dong2021tactilerl}. These sensors observe the topography of the contact surface, which is often a deformable elastomer material to measure contact forces. They also allow for the use of standard visual sensing convolutional neural network architectures since they typically output standard 2D images, making them significantly easier to incorporate into a multi-modal model~\cite{calandra2017feeling}. We explore the ability of the DIGIT~\cite{lambeta2020digit} sensor to improve robotic grasping ability in a simulation environment.

\subsection{Grasping with Tactile sensing}
Tactile information in robotics research has been used in a variety of tactile-relevant applications, which include: tactile exploration, grasping, in-hand manipulation, locomotion, tool manipulation, human-robot interaction, and non-prehensile manipulation~\cite{li2020review}. Although, recent work has shown the strong abilities of analytical approaches to use tactile information, with tasks such as manipulation~\cite{hogan2020tactile}. However, these approaches often rely on assumptions of the geometry of the objects, robot, and environment. On the other hand, learning-based grasping methods do not rely on these assumptions and have gained prominence in recent years for these tasks.

Tactile information has been used in various learning-based approaches, most commonly in supervised learning~\cite{calandra2017feeling,Calandra_2018,veiga2015stabilizing,veiga2018grip,hoelscher2015feature}. Tactile data has also been integrated into reinforcement learning approaches for tasks such as in-hand manipulation~\cite{vanhoof2015learning} and object manipulation~\cite{dong2021tactilerl}. Tactile data has also been used in other learning-based approaches such as unsupervised learning~\cite{cockbum2017stability,tian2019manipulation}, self-supervised learning~\cite{lee2019making}, transfer-learning~\cite{Falco_2019}, and active learning~\cite{driess2017active}. Our work extends previous research~\cite{wang2020tacto} by introducing a variety of real-world object models for predicting simulated grasping outcomes.

\section{Methodology}
\label{sec:experiment}

\subsection{Problem Definition}
Both simulation~\cite{wang2020tacto} and real-world experiments \cite{calandra2017feeling,Calandra_2018} have shown that high-resolution tactile sensing improves the grasp stability estimation for singulated objects compared to more traditional sensing modalities. Our work extends previous simulation studies by using a more complex and standardized benchmark object set designed specifically for robotic grasping.

We assume that a singulated, rigid object is sitting stably on a flat table surface in front of the robot manipulator with a parallel gripper. Further, we assume that a top-down grasp pose is given, either manually or by a grasp detection algorithm, and the robot moves to that grasp pose before closing the gripper. We are interested in solving the problem of estimating whether the grasp would be successful if the object was lifted up without changing the end-effector orientation, given the vision, depth and tactile sensor readings before the lift. The success of a grasp is measured by whether or not the object is above a threshold height after lifting the gripper vertically upwards for a fixed distance and time.



\subsection{Object Models}

\begin{figure}[bt]
\captionsetup[subfigure]{labelformat=empty, justification=centering}

\begin{subfigure}{0.23\columnwidth}
\includegraphics[width=0.9\linewidth]{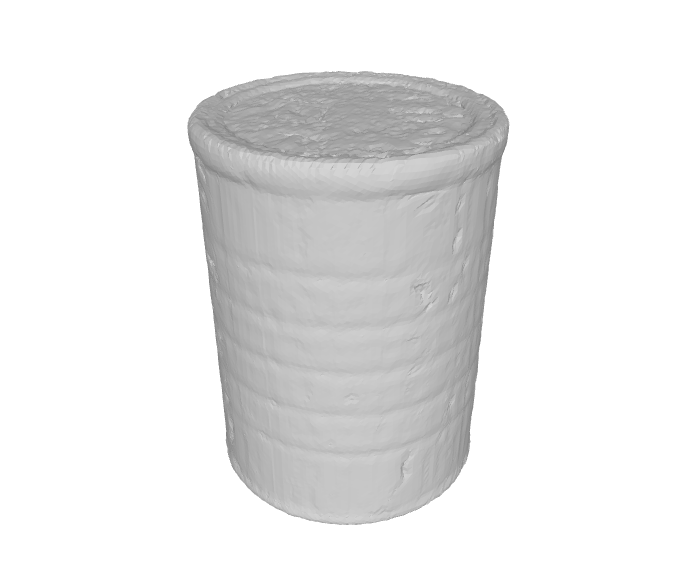}
\caption{Master Chef Can}
\end{subfigure}
\hfill
\begin{subfigure}{0.23\columnwidth}
\includegraphics[width=0.9\linewidth]{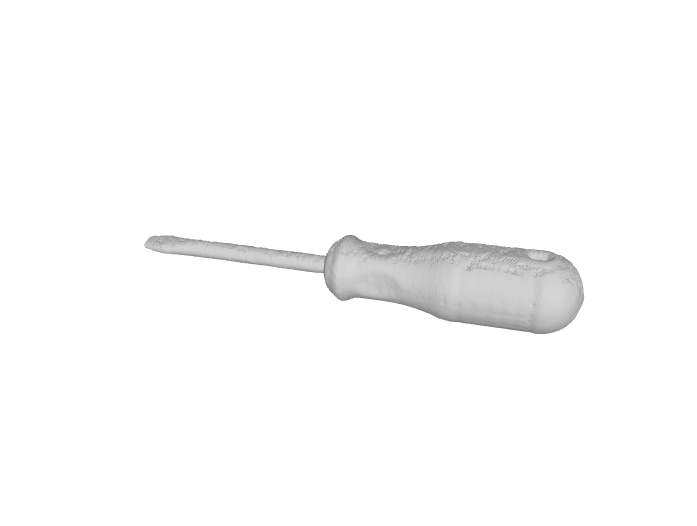}
\caption{Flat Screwdriver}
\end{subfigure}
\hfill
\begin{subfigure}{0.23\columnwidth}
\includegraphics[width=0.9\linewidth]{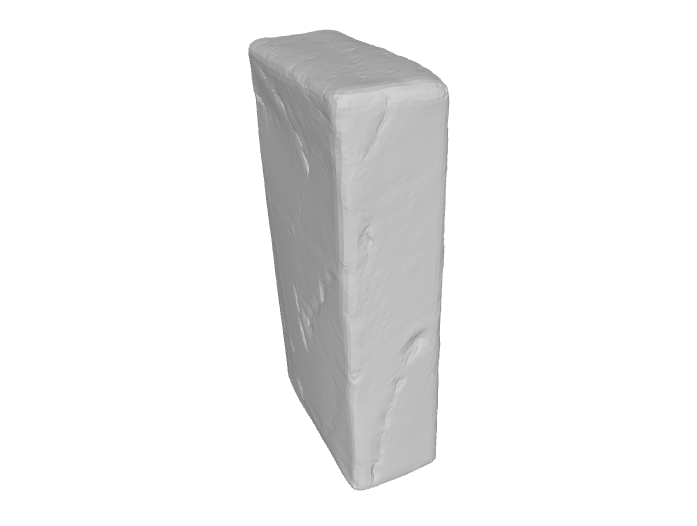}
\caption{Sugar Box}
\end{subfigure}
\hfill
\begin{subfigure}{0.23\columnwidth}
\includegraphics[width=0.9\linewidth]{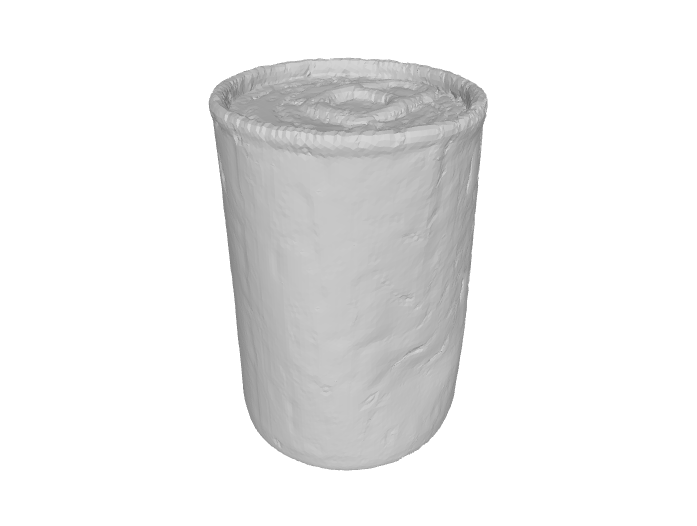}
\caption{Tomato Soup Can}
\end{subfigure}
\hfill

\medskip

\begin{subfigure}{0.23\columnwidth}
\includegraphics[width=0.9\linewidth]{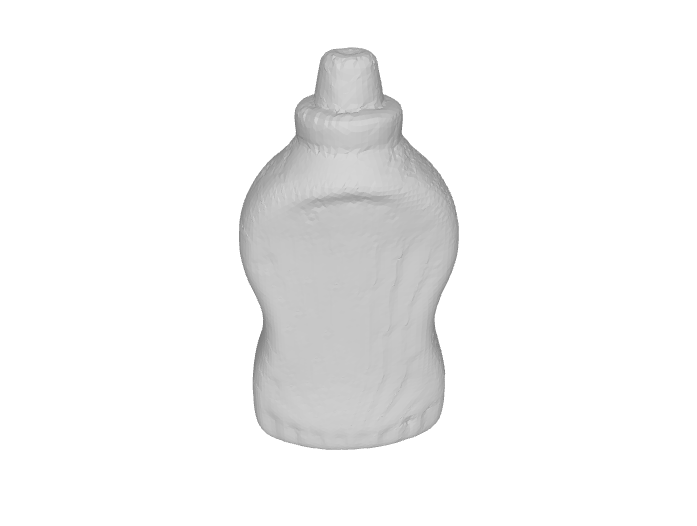}
\caption{Mustard Bottle}
\end{subfigure}
\hfill
\begin{subfigure}{0.23\columnwidth}
\includegraphics[width=0.9\linewidth]{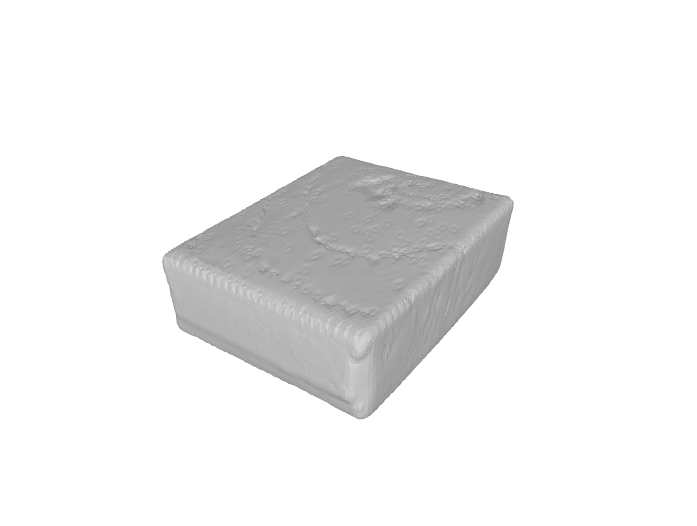}
\caption{Pudding Box}
\end{subfigure}
\hfill
\begin{subfigure}{0.23\columnwidth}
\includegraphics[width=0.9\linewidth]{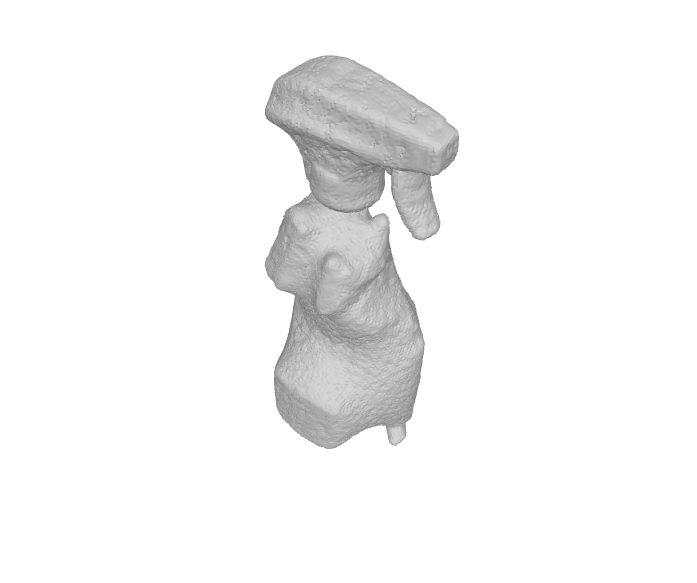}
\caption{Windex Bottle}
\end{subfigure}
\hfill
\begin{subfigure}{0.23\columnwidth}
\includegraphics[width=0.9\linewidth]{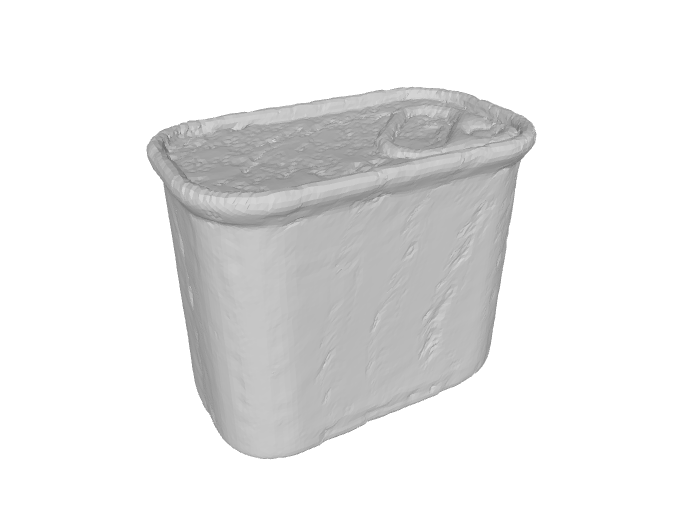}
\caption{Potted Meat Can}
\end{subfigure}

\medskip

\begin{subfigure}{0.23\columnwidth}
\includegraphics[width=0.9\linewidth]{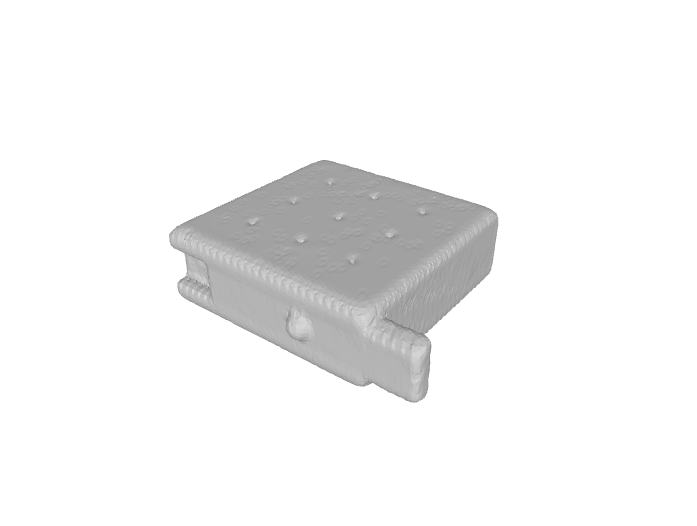}
\caption{Nine Hole Peg Test}
\end{subfigure}
\hfill
\begin{subfigure}{0.23\columnwidth}
\includegraphics[width=0.9\linewidth]{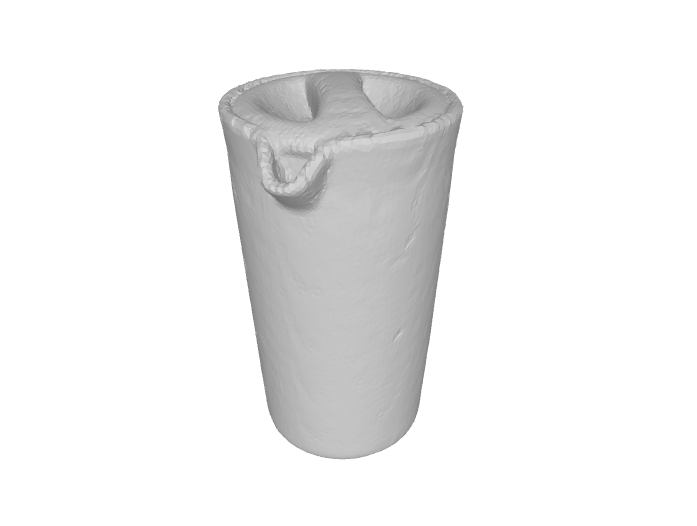}
\caption{Pitcher Base}
\end{subfigure}
\hfill
\begin{subfigure}{0.23\columnwidth}
\includegraphics[width=0.9\linewidth]{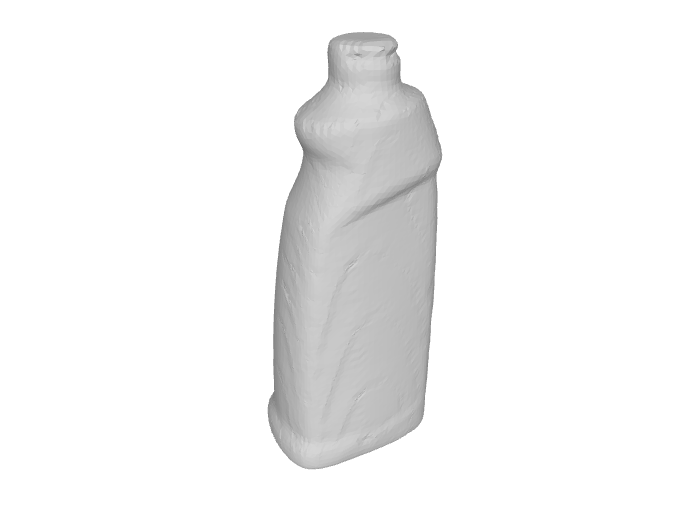}
\caption{Bleach Cleanser}
\end{subfigure}
\hfill
\begin{subfigure}{0.23\columnwidth}
\includegraphics[width=0.9\linewidth]{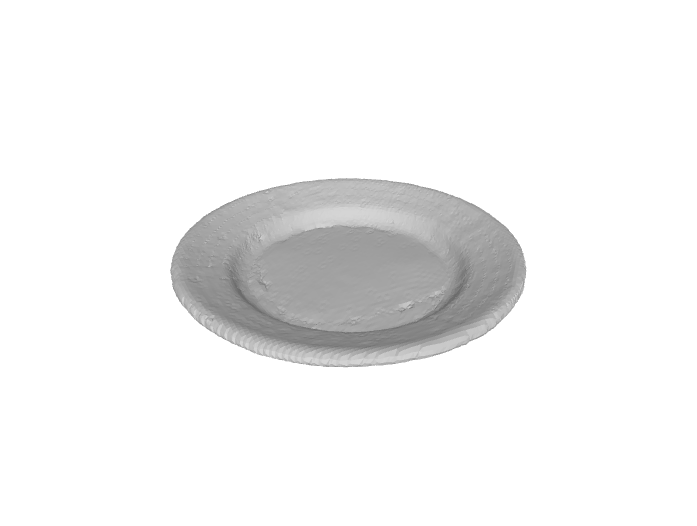}
\caption{Plate}
\end{subfigure}

\medskip

\begin{subfigure}{0.23\columnwidth}
\includegraphics[width=0.9\linewidth]{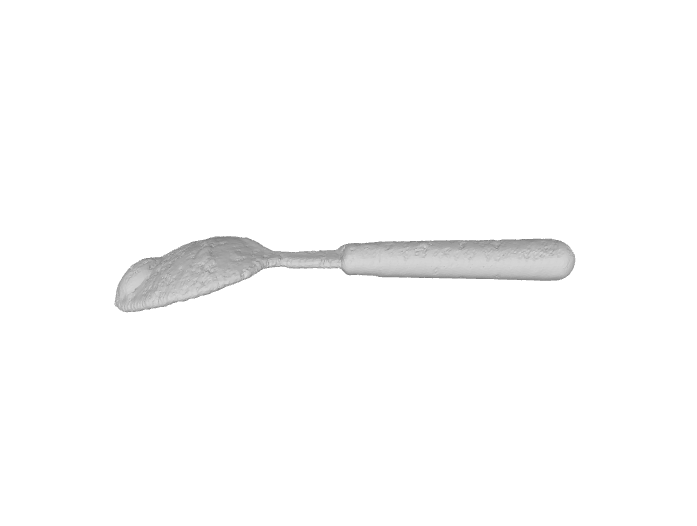}
\caption{Spoon}
\end{subfigure}
\hfill
\begin{subfigure}{0.23\columnwidth}
\includegraphics[width=0.9\linewidth]{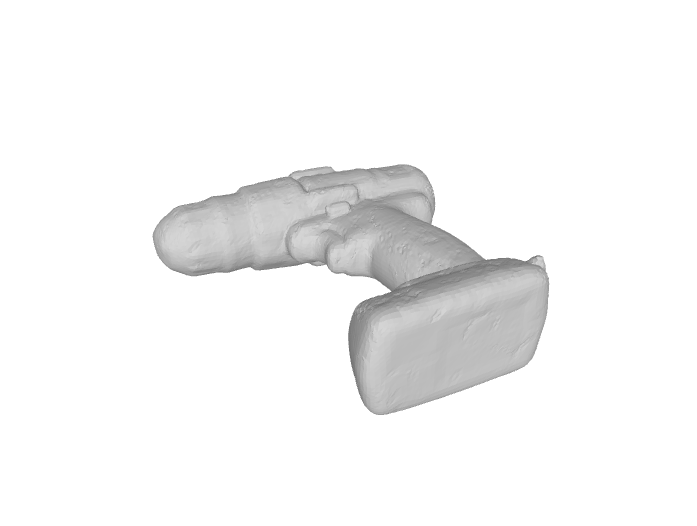}
\caption{Power Drill}
\end{subfigure}
\hfill
\begin{subfigure}{0.23\columnwidth}
\includegraphics[width=0.9\linewidth]{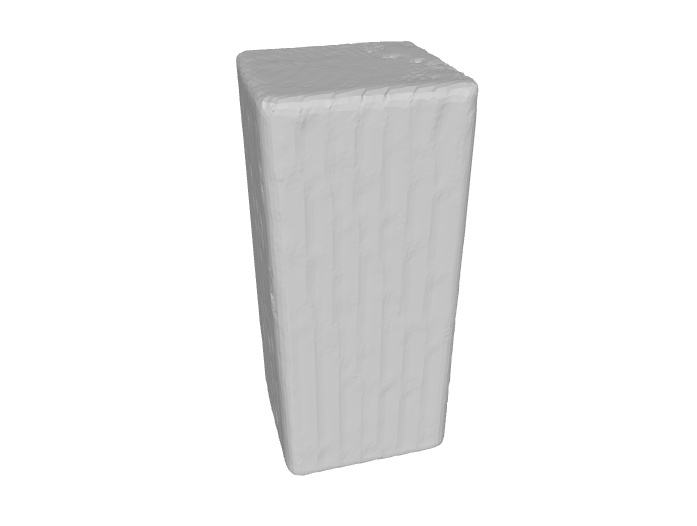}
\caption{Wood Block}
\end{subfigure}
\hfill
\begin{subfigure}{0.23\columnwidth}
\includegraphics[width=0.9\linewidth]{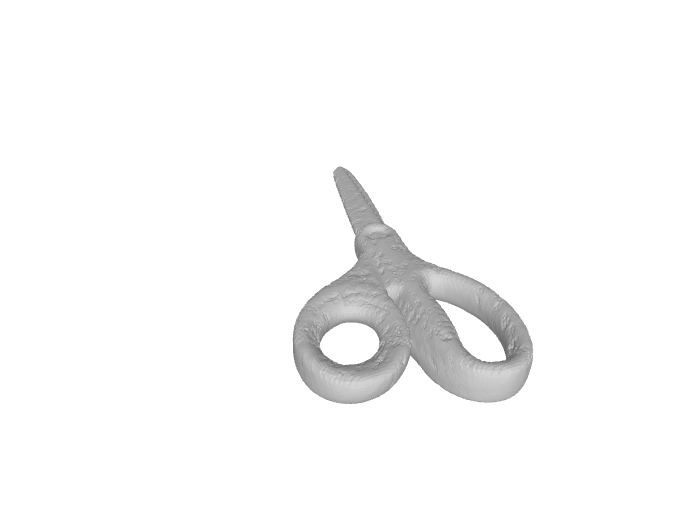}
\caption{Scissors}
\end{subfigure}

\medskip

\begin{subfigure}{0.23\columnwidth}
\includegraphics[width=0.9\linewidth]{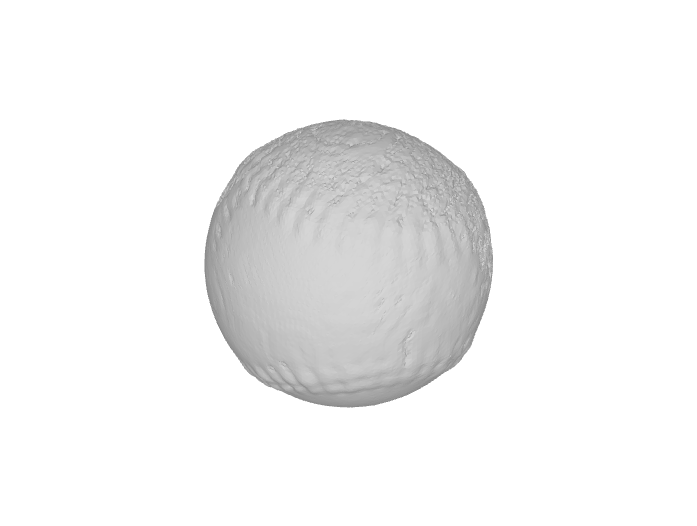}
\caption{Softball}
\end{subfigure}
\hfill
\begin{subfigure}{0.23\columnwidth}
\includegraphics[width=0.9\linewidth]{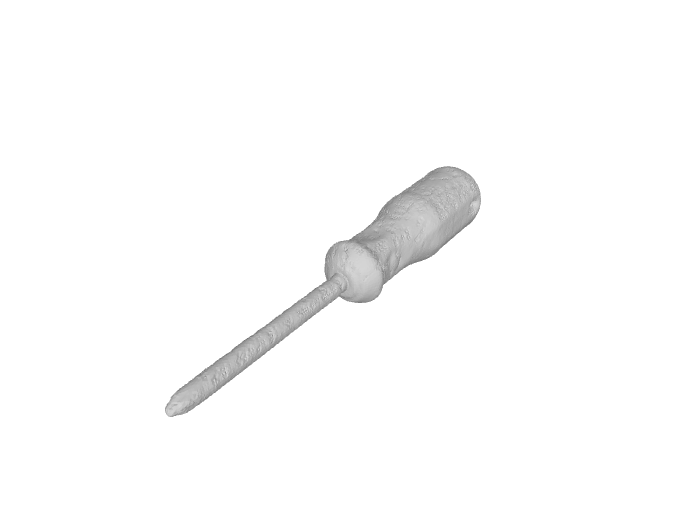}
\caption{Phillips Screwdriver}
\end{subfigure}
\hfill
\begin{subfigure}{0.23\columnwidth}
\includegraphics[width=0.9\linewidth]{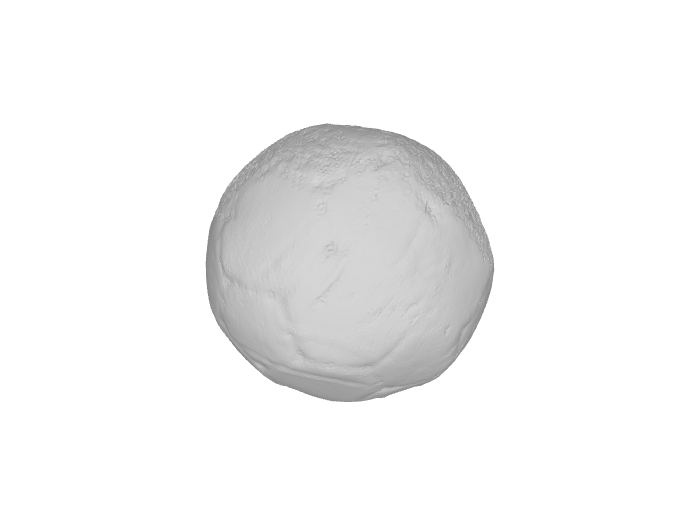}
\caption{Mini Soccer Ball}
\end{subfigure}
\hfill
\begin{subfigure}{0.23\columnwidth}
\includegraphics[width=0.9\linewidth]{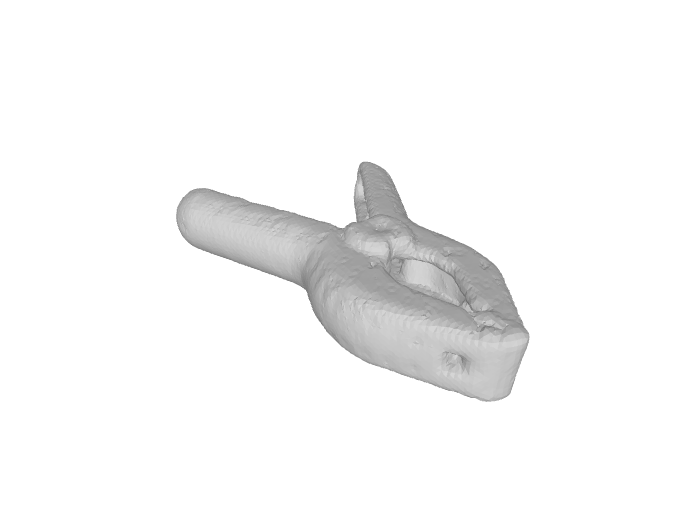}
\caption{Large Clamp}
\end{subfigure}
\caption{20 YCB object models used in training and testing. Watertight mesh equivalents of the object were used to help with the simulation accuracy.}
\label{fig:objects} 
\end{figure}

\begin{figure}[h]
\captionsetup[subfigure]{labelformat=empty, justification=centering}
\begin{subfigure}{0.23\columnwidth}
\includegraphics[width=0.9\linewidth]{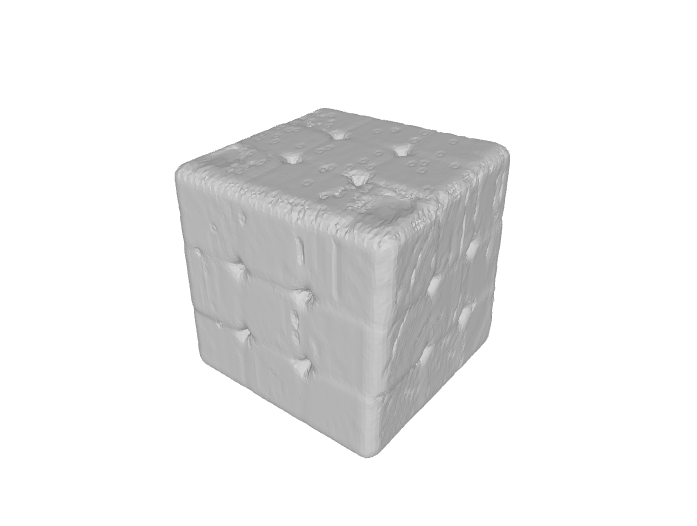}
\caption{Rubik's Cube}
\end{subfigure}
\hfill
\begin{subfigure}{0.23\columnwidth}
\includegraphics[width=0.9\linewidth]{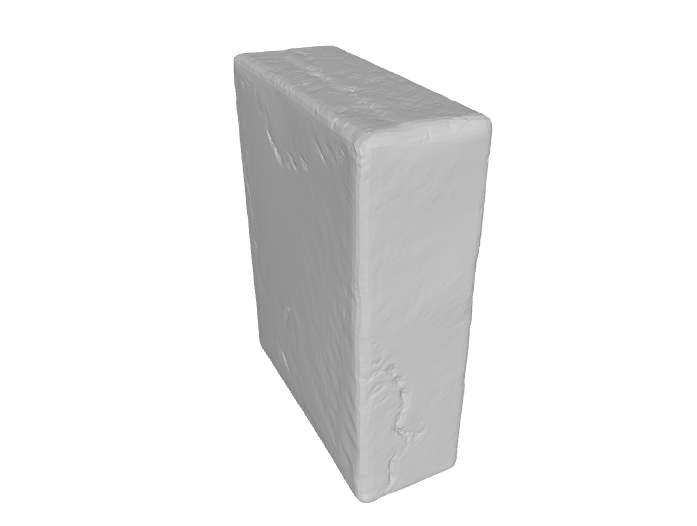}
\caption{Cracker Box}
\end{subfigure}
\hfill
\begin{subfigure}{0.23\columnwidth}
\includegraphics[width=0.9\linewidth]{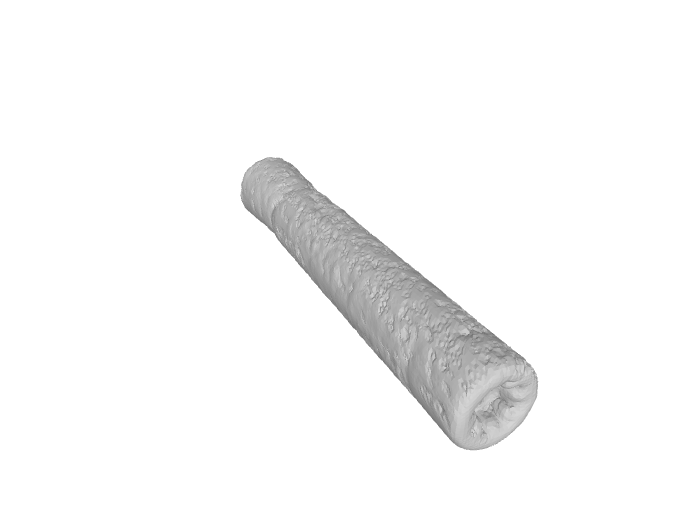}
\caption{Large Marker}
\end{subfigure}
\hfill
\begin{subfigure}{0.23\columnwidth}
\includegraphics[width=0.9\linewidth]{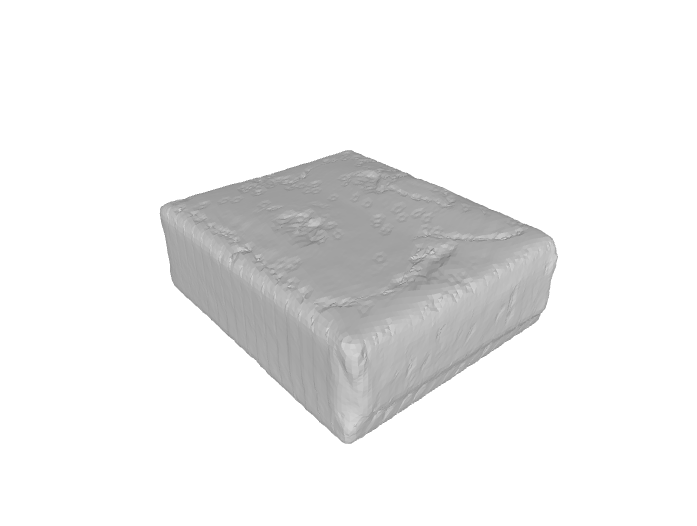}
\caption{Gelatin Box}
\end{subfigure}
\caption{The four YCB object models used as unknown objects in our experiments}
\label{fig:testing_set_object}
\end{figure}

The YCB object set \cite{calli2015ycb} is prominently used as a benchmark in robotic grasping and consists of 77 standard household items such as food, toys, and tools with different shapes, sizes, textures, weight, and rigidity. We use 20 objects from this set for training (shown in Fig.~\ref{fig:objects}) and another 4 objects for validation (shown in Fig.~\ref{fig:testing_set_object}). These objects were selected based on their ability to be grasped successfully by the robotic manipulator, which will be detailed in Sec.~\ref{sec:object_selection}. The YCB dataset provides scanned 3D models of each object obtained from a scanning rig. During preliminary experiments, we observed that grasping physics with these scanned 3D models is often unstable as the gripper moves through the objects during grasp attempts. To improve robustness post-processed watertight 3D models~\cite{corona2020ganhand} of the YCB objects are used. However, no visual textures are available and therefore they were rendered in grey. 

\begin{figure*}[tb]
    \centering
    \includegraphics[width=1\linewidth, trim={3cm 1cm 3cm 1cm},clip]{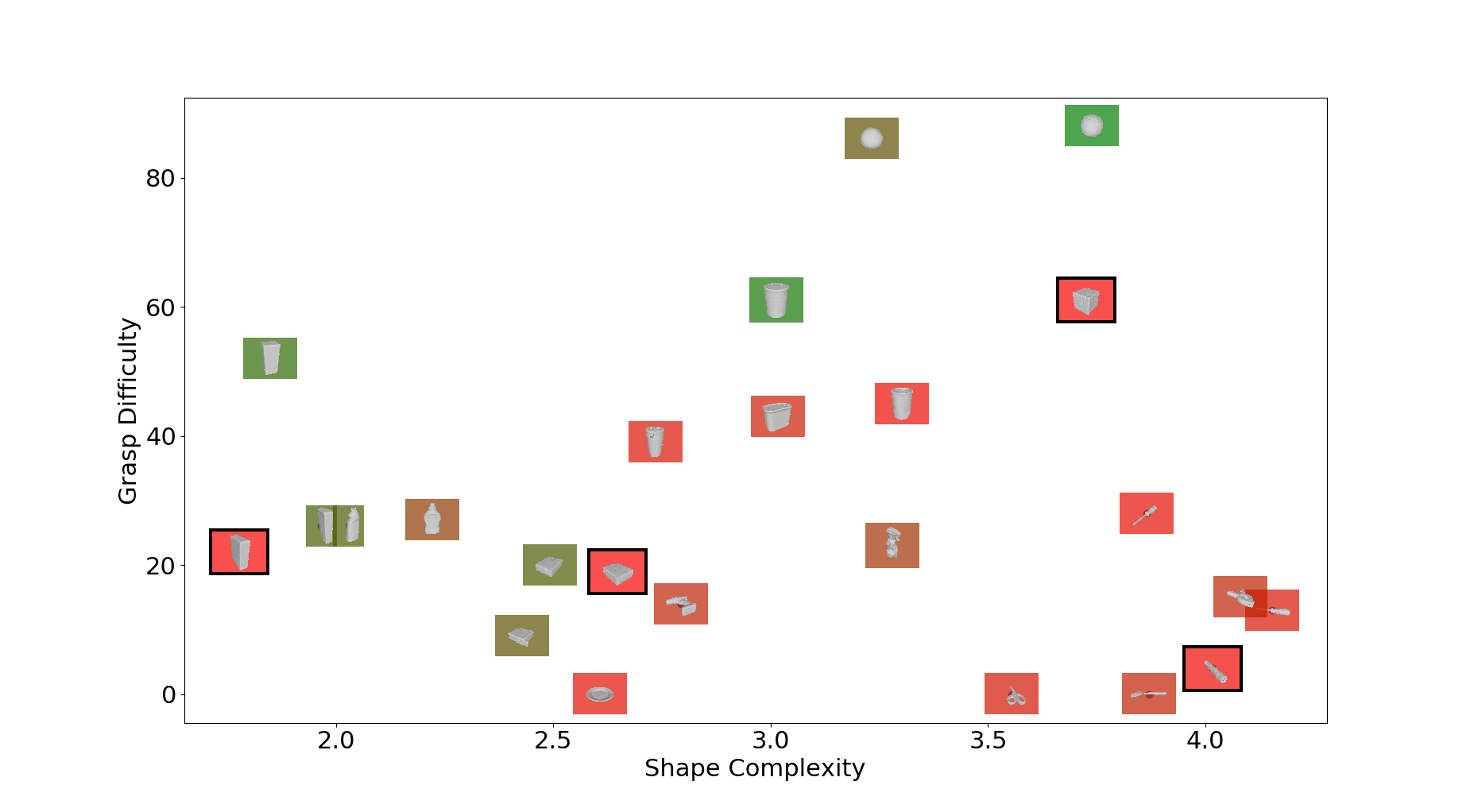}
    \caption{The objects selected to validate grasp predictions plotted by grasp difficulty against shape complexity. The background colour per object indicates the proportion of grasps that contain tactile information on both sensors. From red (low, $\sim3\%$) to green (high, $\sim88\%$).} 
    \label{fig:complexity}
\end{figure*}

Fig.~\ref{fig:complexity} shows the complexity and grasp difficulty of the objects used in our study, as defined by the EGAD! object set~\cite{morrison2020egad}. The grasp difficulty is quantified by the 75th percentile grasp of all sampled grasps, originally proposed by Wang et al.\cite{Wang75}, Dex-Net~\cite{mahler2017dexnet} is used for grasp sampling, and the Ferrari-Canny metric is used for robustness. Morphological complexity is used to measure the complexity of the shape~\cite{complexity1,complexity2}. Objects used for evaluation are shown with the black border, and the color represents the likelihood of collecting good tactile data on a grasp attempt for an object (red is low at approximately. 3\%, green is high at approximately 88\%).



\subsection{Object Selection}
\label{sec:object_selection}

We selected 24 objects among the 77 original YCB objects for our experiments. Object selection occurred in two stages. The first stage involved performing 100 grasp attempts on each watertight object model, scaled by $0.6$ to $0.9$ in increments of $0.1$. This produced a total of 400 grasps per object. We then chose any object with a grasping success greater than 25\% for any individual scaling and selected the scaling with the highest success rate. This is done to eliminate any object models that are inherently too difficult to grasp. This reduced the number of objects to 28. The second stage involved an initial stage of data collection, where any objects which did not produce tactile information on both fingers from 150 sampled grasps were removed. This reduced the number of objects to 24. 

\subsection{Data Collection}
\label{sec:datacollection}
We use PyBullet and TACTO \cite{wang2020tacto} for the simulation. We collected 10,000 data points to make the data comparable to the real-world experiment by Calandra et al.~\cite{calandra2017feeling}. 

A single object is placed on the table at a uniformly random position in the workspace and yaw rotation at each iteration. To collect a data point, we complete three stages of grasping, detailed below.

\begin{enumerate}
\item \textbf{Select grasp pose: } GGCNN~\cite{morrison2018closing} is used to calculate a top-down grasping pose. GGCNN uses a depth image to calculate the quality of grasp at each pixel. We choose the grasp with the highest quality to maximize the grasp success.

\item \textbf{Move and close gripper: }Once the grasp pose is chosen, the robot moves its end-effector to the grasp pose, and the gripper is closed at a constant velocity until at least 2N of force is read on both fingers. Data is only saved if tactile readings on both fingers are detected. This is determined by having at least 100 depth pixels with data greater than 0.0001m. This is because we are interested in both gripper fingers are touching the object, rather than cases where the object has already slipped from the gripper during closing. Furthermore, it would only make sense to try to lift an object in the real-world if both fingers are touching the object. 

\item \textbf{Lift object: }Once the gripper fingers are closed, determined using force feedback, the robot attempts to lift the object vertically upwards. The robot moves at a fixed speed for a fixed amount of time, if the object moves at least 80\% of the vertical distance, the attempt is labeled as successful. 
\end{enumerate}





\subsection{Dataset Filtering}
\label{sec:dataset_filtering}

Simulation of contact is not perfect. For example, the tactile sensors would often not provide information, as fingers were simulated to go through an object. 
The success rate of obtaining tactile input from both fingers differed per object, causing the dataset to have an uneven distribution of datapoints per object. 
The data is filtered to reduce the effect of unbalanced data on the training results. We limit the number of grasp attempts per object to 500 and filtered to create a success/failure split that is as even as possible given collected data.
For example, if an object had 700 grasp attempts, 200 successful and 500 unsuccessful, only 300 unsuccessful would be used for the dataset together with the 200 successful ones. Overall, this filtering provided a dataset with 66.7\% successful grasp labels.

Four of the selected objects presented less than 500 grasps attempts with tactile information. These objects were reserved for the evaluation set.

\begin{figure}[h]
    \centering
    \includegraphics[width=1\linewidth, trim={1cm 1cm 1cm 2cm}]{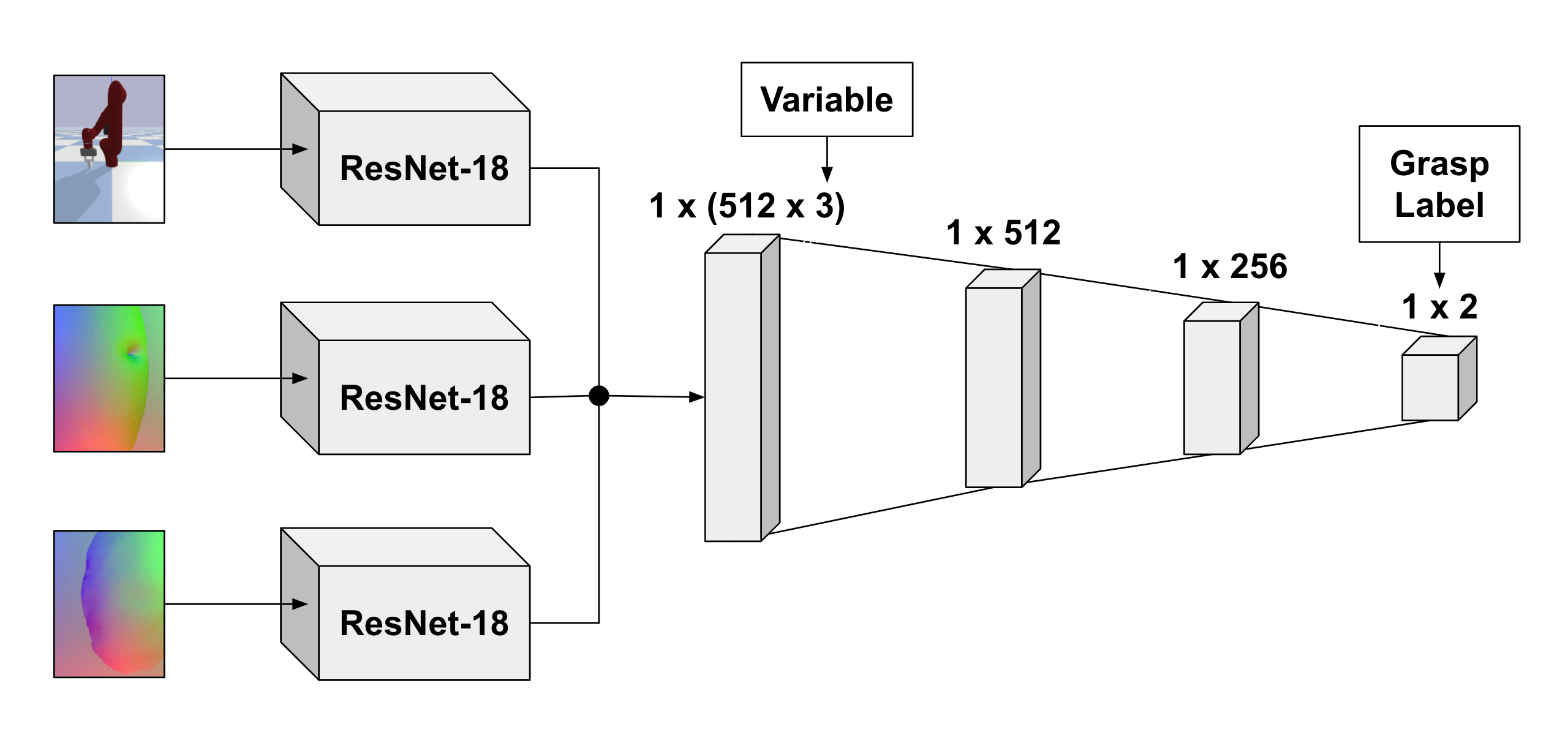}
    \caption{The basic architecture of the networks used to predict grasp success. The inputs to the network were various combinations of sensor modalities.}
    \label{fig:network_diagram}
\end{figure}

\subsection{Training}
Convolutional neural networks with the ResNet-18 backbone~\cite{He_2016_CVPR} were trained using a combination of various modalities, the network architecture is illustrated in Fig.~\ref{fig:network_diagram}. A ResNet-18 backbone was chosen to directly compare it to the real-world experiment by Calandra et al.~\cite{calandra2017feeling}.

In addition to the modalities originally presented~\cite{wang2020tacto}, we added the modalities that included depth (see Table~\ref{tab:test_acc}). In accordance to the original TACTO paper, we trained the networks on $1,000$, $2500$, $5000$ and $10,000$ samples in total (split 80\%/20\% into training and testing). The networks were trained for ten epochs using Binary Cross-Entropy loss, and three-fold cross validation.


\section{Results}
\label{sec:results}
\subsection{Grasp Stability Prediction (Known Objects)}



\subsubsection{Input Sensor Modalities}
\begin{figure}[bt]
    \centering
    \includegraphics[width=1\linewidth, trim={1cm 0cm 1cm 1cm}]{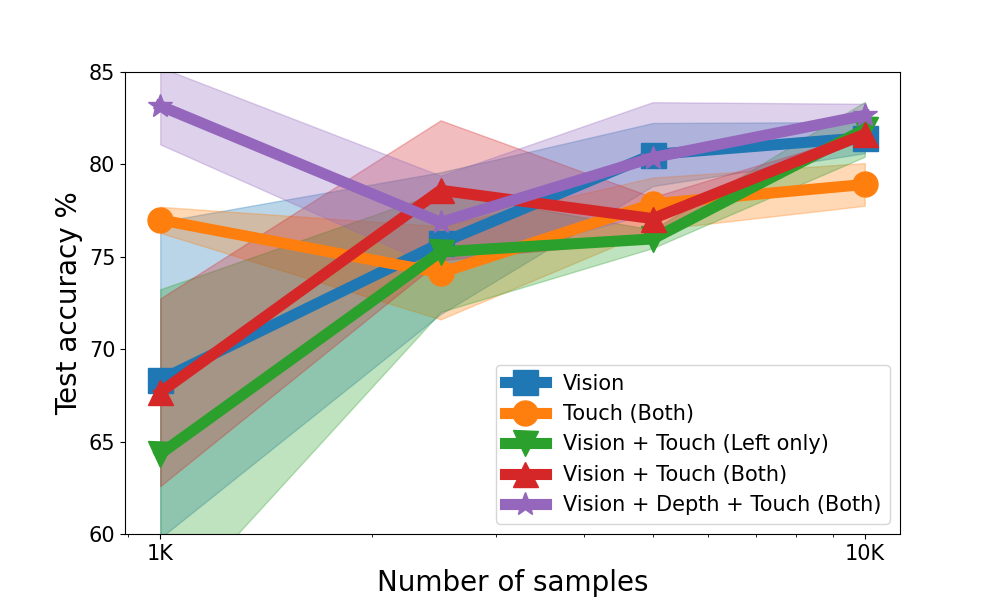}
    \caption{Test accuracy of trained networks on $1,000$, $2,500$, $5,000$ and $10,000$ for various input modalities. The highlighted area bounds the training results from all five cross validation trials.}
    \label{fig:test_acc}
\end{figure}

\begin{table}[bt]
\centering
\begin{tabular} { l |c  }
& Test Accuracy (\%) \\ \hline
Vision + Depth + Touch (Both)&\textbf{82.7±0.6}\\
Vision + Touch (Left only)&81.9±1.5\\
Vision + Touch (Both)&81.6±0.1\\
Vision&81.5±0.8\\
Vision + Depth&80.8±0.4\\
Depth + Touch (Both)&80.8±1.2\\
Depth&80±2.1\\
Touch (Both)&78.9±1.2\\
Touch (Left only)&76±1.5\\ \hline
\end{tabular}
\caption{Ablation study for input sensor modalities. All networks are trained with 10,000 data samples.}
\label{tab:test_acc}
\end{table}

\begin{table}[ht]
\centering
\setlength\tabcolsep{3 pt}
\begin{tabular} { l |c  c  c}
    & \makecell{Vision+\\ Depth+\\ Touch \\ (Both)} & \makecell{Vision+\\ Touch \\ (Left only)}&Vision\\ \hline
Flat Screwdriver&\textbf{68.6±2}&67.1±4.1&68.1±7.7\\
Large Clamp&57.4±10.8&\textbf{59.9±4.4}&57.4±5\\
Spoon&57.6±5.9&56.5±6.3&\textbf{64.1±4.2}\\
Phillips Screwdriver&62.8±2.1&\textbf{66.9±9.3}&60.4±8.8\\
Softball&\textbf{87.5±1.7} & 85.4±6.9&81.2±7.4\\
Scissors&53.1±4.4&\textbf{67.2±6}&59.2±2.1\\
Tomato Soup Can&\textbf{90.3±3.7}&89.5±6.3&85.1±6.7\\
Windex Bottle&81.6±1.6&\textbf{83.8±2}&83.6±4\\
Mini Soccer Ball&92.6±2.7&\textbf{93.6±0.6}&92.3±2.7\\
Potted Meat Can&97.7±0.8&97.9±0.8&\textbf{98.8±0.5}\\
Masterchef Can&\textbf{93.1±5.1}&83.7±2.5&80.8±1.5\\
Power Drill&86±3.6&\textbf{90.8±4.7}&90.1±2.1\\
Pitcher Base&\textbf{91.1±0.7}&88.3±8.4&87.3±5\\
Plate&69.7±1.6&\textbf{79.3±5.7}&74.8±2.6\\
Pudding Box&98.2±2.7&\textbf{98.6±0.9}&97.7±1.1\\
Nine Hole Peg Test&\textbf{84±1.8}&75.4±5.1&78±1.9\\
Mustard Bottle&\textbf{83.8±5.3}&82.9±3.7&80.5±5.4\\
Bleach&\textbf{86.7±4.4}&82.5±1.8&84.6±5\\
Sugar Box&96.8±1.5&96±1.9&\textbf{97.1±0.7}\\
Wood Block&92.3±1.4&90±5.7&\textbf{95±2.1}\\ \hline
\textbf{Average}&\textbf{82.7±0.6}&81.9±1.5&81.5±0.8\\ \hline
\end{tabular}
\caption{Grasp stability prediction accuracy of the 20 ``known objects". The networks are trained on 10,000 data samples. The table is sorted by the shape complexity of the objects (high to low, top to bottom).}
\label{tab:object_acc}
\end{table}

\setlength{\tabcolsep}{0.5mm}
\begin{table}[ht]
\centering
\begin{tabular}{|c|c|c|c|}
\cline{3-4}
\multicolumn{2}{c}{} & \multicolumn{2}{|c|}{Ground Truth} \\
\cline{3-4}
\multicolumn{2}{r|}{} & Successful Grasp & Unsuccessful Grasp \\
\hline
\multirow{2}{*}{\rotatebox[origin=r]{90}{Network Output}} & 
\raisebox{0cm}{\rotatebox{90}{\footnotesize Successful Grasp}} & {\includegraphics[width=0.44\linewidth]{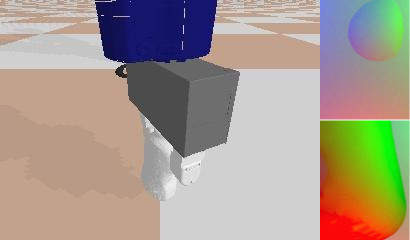}} & {\includegraphics[width=0.44\linewidth]{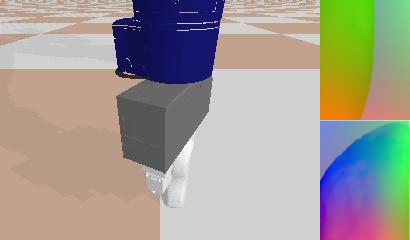}} \\ 
\cline{2-4} & \raisebox{0cm}{\rotatebox{90}{\scriptsize Unsuccessful Grasp}} & {\includegraphics[width=0.44\linewidth]{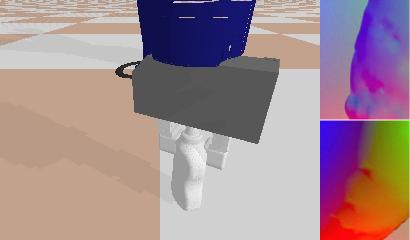}} & 
{\includegraphics[width=0.44\linewidth]{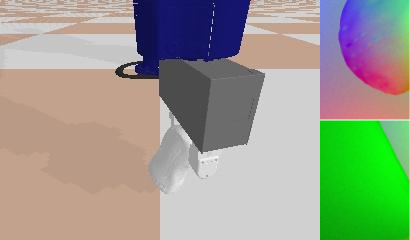}} \\ \hline
\end{tabular}
\caption{Vision + Depth + Touch (Both) model predicted grasp success against the ground truth for four samples on the mustard bottle object.}
\label{tab:successandfailures}
\end{table}
\setlength{\tabcolsep}{0.15cm}

\begin{table*}[ht]
\centering
\begin{tabular} { l | c c c c | c}
& Rubik's Cube & Cracker Box & Large Marker & Gelatin Box & Average\\ \hline
Vision + Depth + Touch (Both)&49.1±0.6&87.8±4.7&50.4±4.7&47.8±4.0&58.0±3.5\\
Vision + Touch (Left only)&49.9±1.4&79.8±10.8&47.0±6.0&41.1±7.8&53.9±1.9\\
Vision + Touch (Both)&48.6±6.4&88.9±7.5&48.9±7.8&51.9±12.1&58.5±3.8\\
Vision&48.9±0.5&81.2±3.7&42.9±5.4&32.9±4.1&51.1±2.7\\
Vision + Depth&\textbf{52.8±2.0}&69.6±7.4&39.8±2.5&34.1±3.1&48.5±1\\
Depth + Touch (Both)&45.3±2.8&\textbf{93.5±0.9}&56.4±2.0&52.9±10.6&\textbf{61.3±2.8}\\
Depth&44.6±1.2&79.5±13.4&41.3±5.9&42.1±1.0&51.6±4.9\\
Touch (Both)&44.2±1.7&93.3±1.9&55.8±4.2&\textbf{53.5±5.5}&60.9±3.1\\
Touch (Left only)&47.7±6.2&86.2±5.5&\textbf{58.2±5.5}&51±11.1&60.3±4.4\\
\hline
\end{tabular}
\caption{Grasp stability prediction accuracy on unknown objects for networks trained on 10,000 samples. The overall performance is much lower than for known objects.}
\label{tab:evaluation}
\end{table*}

We perform an ablation study for a combination of input modalities, results are shown in Fig.~\ref{fig:test_acc} and Tab.~\ref{tab:test_acc}. 
Our results suggest that using all sensors (vision, depth, and touch) led to the highest accuracy (Fig.~\ref{fig:test_acc}), albeit with a narrow margin with larger number of samples, as there was only 6.7 percentage points between the best- and worst-performing models for 10,000 samples. We note that the accuracy didn't necessarily increase with the number of samples for lower sample sizes. However this can likely be explained by the higher variance present for lower sample sizes as the test set size scaled with the training set.

The prediction accuracy for each modality increases with the number of training samples provided. Moreover, this convergence, particularly of vision, appears to happen significantly faster than previous grasping work using the TACTO simulator~\cite{wang2020tacto}.

While the different modalities converge when 10,000 samples are used (Table~\ref{tab:test_acc}), integrating touch with visual and depth sensing provides significant improvement when the data availability is limited (83.5\% when trained with 1,000 samples).


Another interesting observation was that the networks that utilized input from the left tactile sensor alone performed just as well as the corresponding networks which used input from both tactile sensors. This result is likely a byproduct of the filtering process during data collection (see Sec.~\ref{sec:dataset_filtering}), which ensured tactile readings from both sensors. Another possible factor is that fewer input images were easier to train as there are fewer network weights to learn. This observation suggests that a single high-resolution vision-based tactile sensor could be sufficient when combined with a potentially cheaper tactile sensor that only detects whether the gripper finger has made contact with the object. 


\subsubsection{Performance on individual objects}

The grasp stability prediction accuracy of the top-performing networks on the objects used in training are detailed in Table~\ref{tab:object_acc}. The objects are ordered in descending levels of complexity. We notice a large variation among objects in terms the grasp stability accuracy: for the highest performing modality (all sensors combined), the Pudding Box object had 98\% accuracy, whereas the Scissors object had only 53\% accuracy.

Our results show that, overall, the inclusion of tactile information marginally increases overall network performance on known objects. Additionally, networks that include the touch modality tend to perform better on objects of higher complexity. Furthermore, there tends to be an advantage of using tactile information for thinner objects, such as the two Screwdriver objects and the Scissors object. However, this is not always true, as evident by the vision-only model performing best on the Spoon, Potted Meat Can, Sugar Box and Wood Block objects. We notice a trend that larger box shaped objects tended to have higher accuracy for the vision-only network, but we have no explicit explanation for why this is the case. A further trend was for thin objects (e.g Spoon, Screwdrivers, Large Clamp, Scissors) to have relatively low accuracy. This could be because there was greater difficulty in predicting the outcome, resulting in inconsistent results over different modalities. 



\subsubsection{Qualitative Analysis}
Table \ref{tab:successandfailures} illustrates four examples of successful and unsuccessful predictions of the network for the Vision + Depth + Touch (Both) model. 

The top left image illustrates a correctly predicted successful grasp which is likely due to its clear vision of the object and centered grasp. 

A common failure of the network is due to partial occlusion of the object caused by the gripper. An example of this can be seen in the top right image resulting in an unsuccessful grasp being incorrectly predicted.


The bottom left image depicts the network incorrectly predicting that the grasp would be unsuccessful. Likely, this example could also be suffering from vision occlusion of the point contacts by the gripper.

The bottom right image of a correctly predicted unsuccessful grasp illustrates the network's potential ability to recognize a falling object and identify the low chance of a successful grasp.

These examples support our previous finding that the vision modality is heavily relied on for grasp prediction accuracy. This observation indicates perhaps more work is needed to rely more on tactile information in poor visual situations.

\subsection{Grasp Stability Prediction (Unknown Objects)}


The results on unknown objects in Table~\ref{tab:evaluation} reveal that the best performing networks on the known objects did not align with the highest performing networks on unknown objects. This illustrates an overall issue around generalization and the potential the networks over-fitted to the training data. This is particularly evident by the poor results on the "Gelatin Box" (48\%) compared to the very similarly shaped "Pudding Box" (98\%). However, the networks that utilized tactile information did generalize better -- the three worst-performing networks did not incorporate tactile information.


\section{Discussion}
\label{sec:discussion}

\textbf{Best sensing modalities:} Combining  Vision + Depth + Touch (Both) modalities created the most successful predictions for known objects. It also provided high levels with only limited data (1,000 samples instead of 10,000). The same modality showed third-best performance for unknown objects. This result indicates that there is no clear winner in terms of which sensor modality should be utilized for grasp stability prediction. 
Additionally, this result may be subject to over-fitting of the networks to the training objects as suggested in Table~\ref{tab:evaluation}. This could be a result of a lack of data augmentation to allow the network to generalise.

For known objects, the three worst-performing modalities do not include vision illustrating the criticality of vision for predicting grasp success. There is further evidence on this observation, as we noticed that in many cases where the network was not able to predict grasp success, the object was occluded by the gripper. This problem could be alleviated with a multi-camera or an eye-in-hand setup.


\textbf{Comparison to existing literature:} Our best performing modality on known objects, Vision + Depth + Touch (Both), was actually not tested by Calandra et al.\cite{calandra2017feeling}, who did not explore depth sensing in detail. \cite{calandra2017feeling}'s best performing modality was Touch + Vision, however our work suggests that adding depth could further improve their results.

We observe that the performance of different modalities in our work was much closer to each other compared to previous work on real robots~\cite{calandra2017feeling} and in the TACTO simulator~\cite{wang2020tacto}. However, we did observe a similar trend of performance increasing with the number of modalities utilized, confirming this previous work that the addition of sensing modalities can improve grasp prediction performance.

\textbf{Problems with the simulator:} We encountered several difficulties during the data collection phase. When attempting to grasp the object, the gripper would often pass through the object. This occurred very frequently when we used the original 3D scans from the YCB dataset. This was significantly reduced after switching to the watertight models, however, these cases still happened, therefore, faulty data needed to be filtered out. Furthermore, we often encountered successful grasps with no tactile information on one or both DIGIT sensors and no force feedback information. As such, we needed to perform significant filtering of our data to remove such cases. This extensive filtering left approximately 20\% of the original data collected, notably increasing the time required for data collection. As well, due to this extensive filtering, there is a possibility that this may bias results as certain grasp types may be more likely to be included.

\textbf{Object diversity: }As depicted in Fig.~\ref{fig:complexity}, we observe that shapes of higher complexity tended to be more difficult to obtain grasping data with valid tactile information. However, this was not true for grasp difficulty, which did not significantly affect on the amount of data that could be collected. This illustrates a barrier in the simulator of dealing with objects of higher complexity, but not grasp difficulty.

\section{Conclusion and Future Work}
\label{sec:conclusion}
The movement towards learning-based approaches in robotic grasping has increased the need for efficient large-scale data collection and subsequently, the ability to accurately simulate  sensors during robotic grasping. In this paper, we investigate the use of high-resolution tactile sensor information in a data-driven approach to estimate the success of robotic grasping.

Building on previous work~\cite{calandra2017feeling} in a simulation setting, we used a subset of objects from the YCB benchmark. We found evidence that using multiple modalities helped with predicting whether a grasp would be successful. 
Our attempt at training and performing robotic grasping of benchmark objects in the TACTO simulator revealed several points of difficulty, particularly in data collection. For example, the simulator requires watertight object models and filtering to remove data inconsistencies, especially to ensure tactile information is present.

Our results suggest that a combination of visual, depth, and tactile sensing provides good predictions for known objects, especially in data limited scenarios. 
We also see, in our findings, the possibility of using a single vision-based tactile sensor with a force/torque sensor could provide the sufficient tactile information needed to predict grasp success on par with using two vision-based tactile sensors. This could be an interesting direction for future work. Additionally, we observed many failure cases occurred in visually occluded settings, and as such future work could pursue the ability to rely on tactile information more heavily in these situations.

\bibliographystyle{IEEEtran}
\bibliography{IEEEabrv,crv}

\end{document}